\pdfoutput=1

\documentclass[11pt]{article}

\usepackage[]{acl}

\usepackage{times}
\usepackage{latexsym}
\usepackage{graphicx}
\usepackage{amsmath}
\usepackage{bbm}
\usepackage{subcaption}
\usepackage{geometry}
\usepackage{multirow}
\usepackage{array}
\usepackage{makecell}
\usepackage{hhline}
\usepackage{booktabs}
\usepackage{float}

\usepackage[T1]{fontenc}

\usepackage[utf8]{inputenc}

\usepackage{microtype}

\title{WaterJudge: Quality-Detection Trade-off when Watermarking Large Language Models}

\author{Piotr Molenda, Adian Liusie, Mark J. F. Gales \\
  ALTA Institute, Department of Engineering, University of Cambridge \\
  \texttt{pm725@cam.ac.uk, al826@cam.ac.uk, mjfg@eng.cam.ac.uk} \\}
\begin{document}
\maketitle
\begin{abstract}

Watermarking generative-AI systems, such as LLMs, has gained considerable interest, driven by their enhanced capabilities across a wide range of tasks. Although current approaches have demonstrated that small, context-dependent shifts in the word distributions can be used to apply and detect watermarks, there has been little work in analyzing the impact that these perturbations have on the quality of generated texts. Balancing high detectability with minimal performance degradation is crucial in terms of selecting the appropriate watermarking setting; therefore this paper proposes a simple analysis framework where comparative assessment, a flexible NLG evaluation framework, is used to assess the quality degradation caused by a particular watermark setting. We demonstrate that our framework provides easy visualization of the quality-detection trade-off of watermark settings, enabling a simple solution to find an LLM watermark operating point that provides a well-balanced performance. This approach is applied to two different summarization systems and a translation system, enabling cross-model analysis for a task, and cross-task analysis.

\end{abstract}

\section{Introduction}

Large Language Models (LLMs) have progressed tremendously and are capable of generating high-quality texts for a diverse range of tasks. While these systems enhance automation, concerns arise about potential misuse, such as students using chat assistants for assignments or malicious users generating fake news articles. To counter this, current work has introduced the idea of LLM watermarking \cite{kirchenbauer2023watermark}, where imperceptible patterns are injected into the generated text, enabling the statistical identification of whether text was generated by an LLM or not. However, most proposed watermarking schemes restrict the output generation space, which may lead to a trade-off between quality and watermarking detection performance. Although there has been great effort into improving watermarking schemes for LLMs \cite{yoo2023robust, kuditipudi2023robust, kirchenbauer2023reliability}, less work has analyzed the resulting quality degradation. It is common for watermarking schemes to measure quality by reporting the perplexity from a larger pre-trained LLM \cite{kirchenbauer2023watermark, takezawa2023necessary, wang2023towards, zhao2023provable, ren2023robust, liu2023semantic}, or to report similarity metrics such as BLEU or ROUGE\cite{fu2023watermarking, takezawa2023necessary, li2023improving, kirchenbauer2023reliability}, however, these metrics are simplistic heuristics and may not truly capture actual output text quality, as discussed by \citet{zhong2022towards, wang2022perplexity, zheng2023judging}. 

\begin{figure}[t]
    \centering
    \includegraphics[width=\columnwidth]{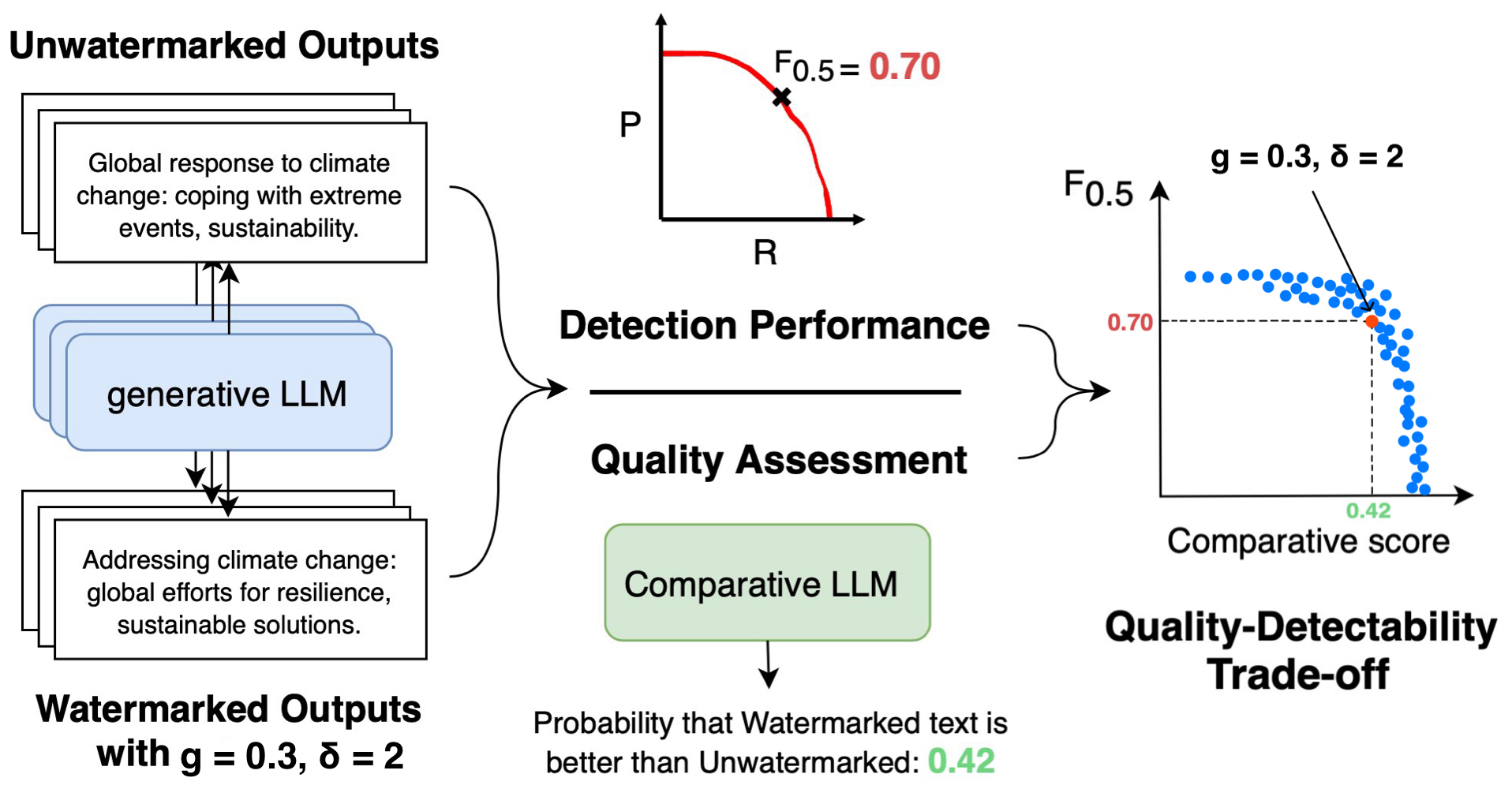}
    \caption{High-level overview of the WaterJudge Framework: Given a system, watermarking parameters, and set of inputs, watermarked outputs are assessed in terms of quality and detectability, leading to a curve over all operating points.}
    \label{fig:diag_1}
    \vspace{-5mm}
\end{figure}

This work proposes WaterJudge, a framework for analyzing the trade-off between watermarking detectability and the quality of generated watermarked text. We leverage the LLM-as-a-judge evaluation approaches \cite{zheng2023judging, liusie2023zero} to measure the average probability that an LLM prefers a watermarked text over an unwatermarked text. This is used as a metric for quantifying the quality degradation caused by watermarking, which with watermark detection performance, can be used to determine the quality and detectability of a watermark operating point. This provides an approach for practitioners to visualize the effectiveness of specific watermarking operating points, enabling simple selection of an optimal watermark setting with minimal quality degradation.

\section{WaterJudge}
\label{sec:theory}
\vspace{-1mm}

\subsection{Soft-Watermarking Scheme}
\label{sec:simple_wm}
Language Models predict the conditional distribution of the next token $w_{i+1}\!\in\!\mathcal{V}$ given the input text $x_{1:M}\!\in\!\mathcal{V}^{M}$ and the previously generated tokens, $w_{1:i}$. For identification of LLM generated text, \citet{kirchenbauer2023watermark} propose a simple soft-watermarking scheme, where the previous token $w_{i}$ is used in a hash function to split the vocabulary into a mutually exclusive green list $\mathcal{V}_{\tt g}(w)$ and red list $\mathcal{V}_{\tt r}(w)$. The approach then incentivizes green-list words to be generated at the next step, such that the green-list word count can determine whether a text was generated by the LLM or not. The parameter $g$ sets the relative size of the green list, such that $|\mathcal{V}_{\tt g}|$ = $g \cdot |\mathcal{V}|$ and $|\mathcal{V}_{\tt r}| = (1\!-\!g) \cdot |\mathcal{V}|$. The watermarking scheme then increases the logits of all tokens in the green list by a bias $\delta$,
\begin{equation}
    l^{wm}_{k} = 
    \begin{cases}
        l^{lm}_{k} + \delta, & \text{if } w^{lm}_{k} \in \mathcal{V}_{\tt g} \\
        l^{lm}_{k}, & \text{otherwise}
    \end{cases}
\end{equation}



\noindent Where $l_k$ is the logit for the k'th token in the vocabulary $w^{lm}_{k}$. The watermarking scheme therefore has two parameters, the green list size $s$ and the green list bias $\delta$. The watermark score for a particular text is then calculated as the number of green list words present in the output text, where the higher the score, the more likely the output was generated by the watermarked LLM. 
\begin{equation}
    s_{\tt wm} = \frac{1}{N_w} \sum_{i=1}^{N_w} \mathbbm{1}(w_i \in \mathcal{V}_{\tt g}(w_{i-1}))
\end{equation}
\noindent Where $\mathbbm{1}$ is the indicator function. Note that this watermarking scheme has the useful property that detection can be achieved even without model access. One only requires knowledge of the tokenizer and hashing function as this enables the green and red lists to be dynamically calculated, which is all that's needed to score texts. Further, if multiple models share a tokenizer, there could be an agreed watermarking convention that enables universal watermark detection over a range of models.


\subsection{Zero-shot Comparative Assessment}
\label{sec:zero_shot_comp}
LLM comparative assessment \cite{liusie2023zero, zheng2023judging}, which prompts an LLM to determine which of two texts is better, is used in our framework to measure the quality degradation caused by watermarking. This method was selected due to being simple, zero-shot, and easily transferable to a range of tasks, as well as demonstrating impressive NLG evaluation performance. 

\begin{figure}[t]
    \centering
    \includegraphics[width=\columnwidth]{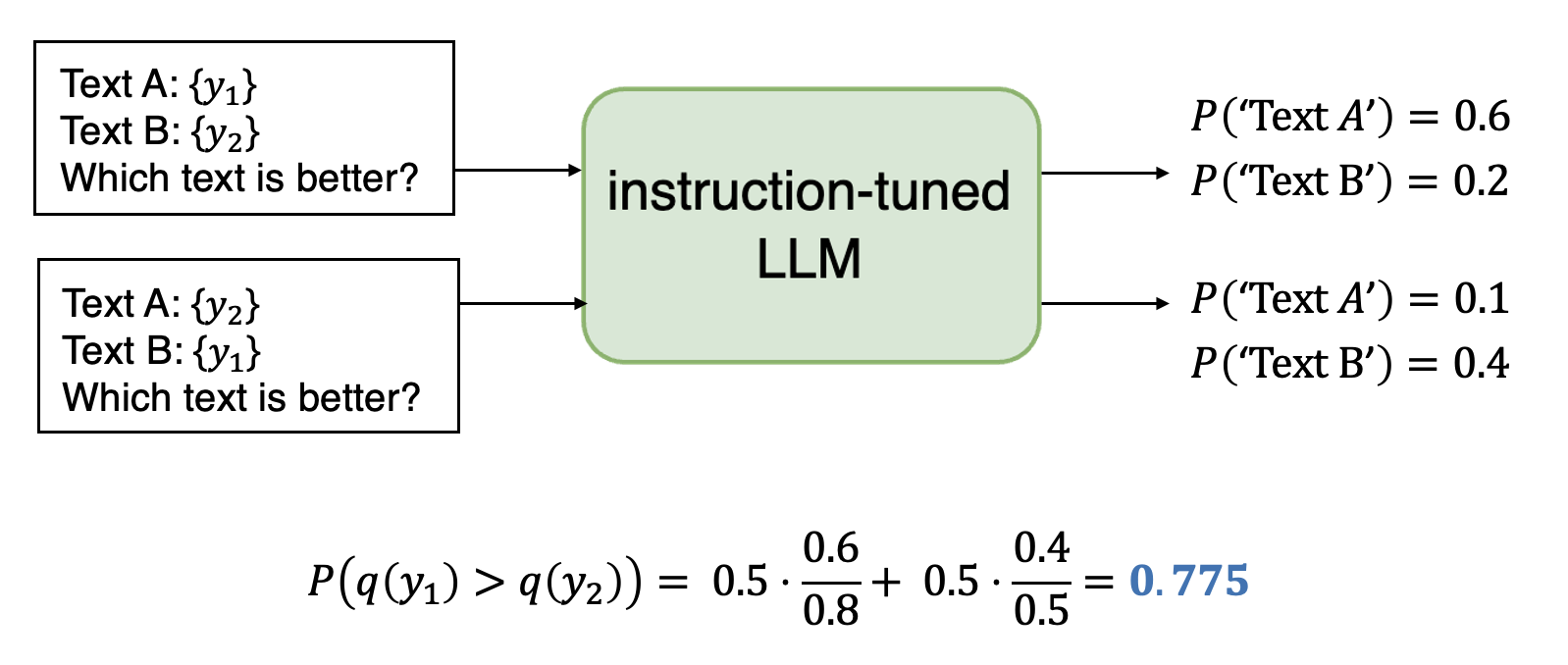}
    \caption{Comparative assessment probabilities are attained by calculating the likelihood of generating `Text A' or `Text B', normalizing, and averaging over both permutations. Example prompts are displayed, with the actual prompts shown in Appendix \ref{sec:appendix_prompts}.}
    \label{fig:comparative_diagram}
    \vspace{-4mm}
\end{figure}

For a given task and model, let $x$ represent the input text, $y$ the generated output text, and $y_{\tt wm}$ an output text generated from the system when watermarked. The Comparative assessment uses open-sourced instruction-tuned LLMs by querying which of the two provided texts is better. The comparative assessment system outputs $P(q(y_1)\!>\!q(y_2)|x)$, the probability that the quality of text $y_1$ is better than the text $y_2$, as demonstrated in Figure \ref{fig:comparative_diagram}. The watermark degradation is measured over a corpus of input texts $\mathcal{D} = \{x^{(i)} \}_{i=1...N_d}$, with the average comparative selective probability used as the quality metric \vspace{-4mm}
\begin{equation}
    s_{\tt q} = \frac{1}{N_d} \sum_{i=1}^{N_d} P(q(y_{\tt wm}^{(i)}) > q(y^{(i)}) | x^{(i)})
\end{equation} 

\vspace{-1mm}

\noindent where $y^{(i)}$ and $y_{\tt wm}^{(i)}$ are the generated base and watermarked outputs respectively, given input $x^{(i)}$.

\begin{figure*}[t]
     \centering
     \begin{subfigure}{0.42\textwidth}
        \includegraphics[width=\linewidth]{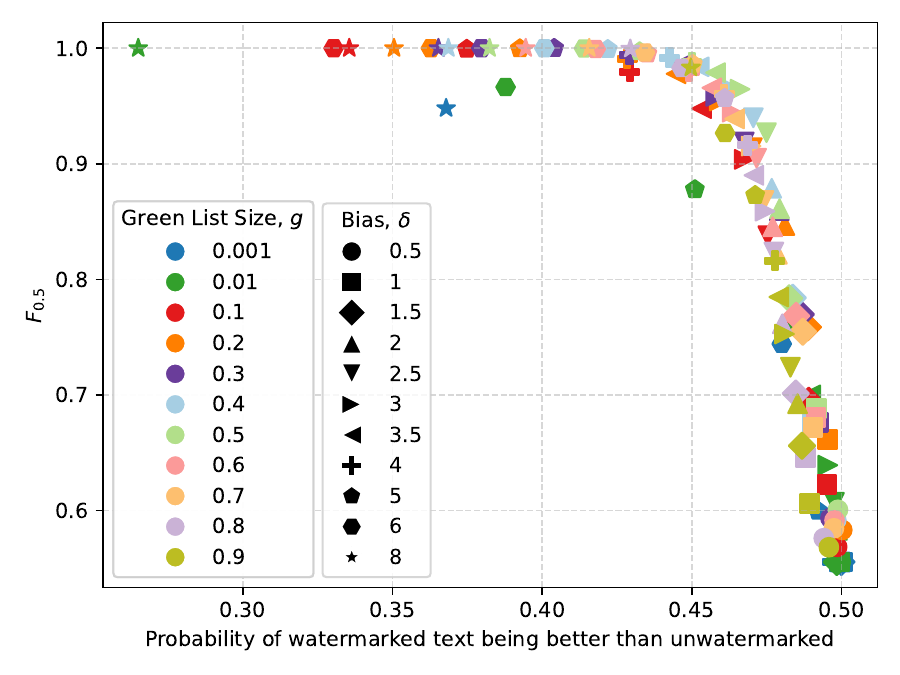}
        \caption{BART}
        \label{fig:bart}
     \end{subfigure}
     \hspace{0.05\textwidth}
     \begin{subfigure}{0.42\textwidth}
         \includegraphics[width=\linewidth]{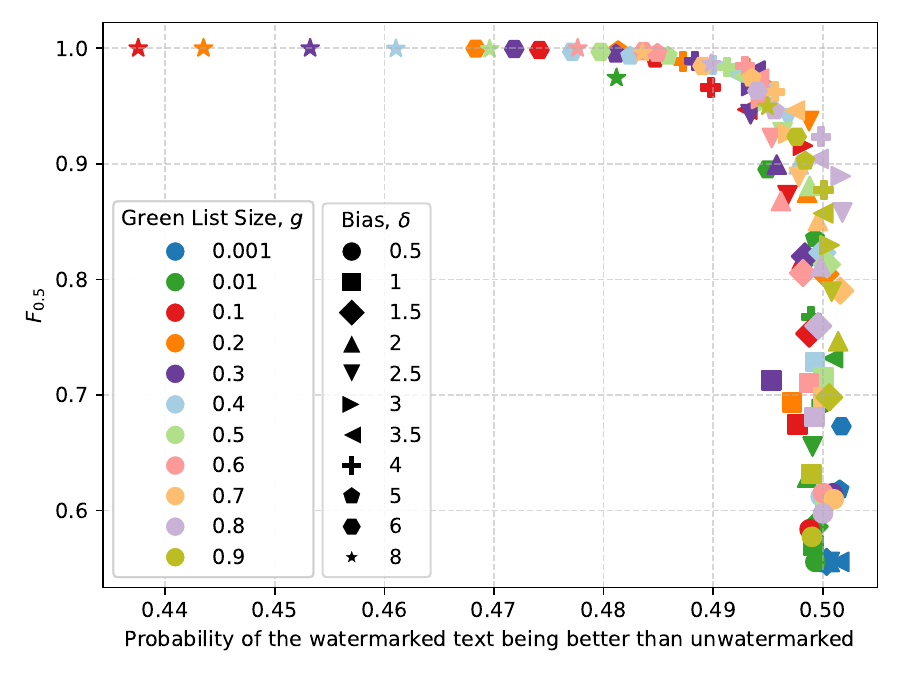}
        \caption{Zephyr}
         \label{fig:zephyr}
     \end{subfigure}
     \vspace{-3mm}
     \caption{The trade-off between quality and detectability when watermarking. Each point is a watermark setting with green list size $g$ and bias $\delta$, displaying $F_{0.5}$ detectability score and average Comparative Assessment probability.}
     \label{fig:appendix_zephyr}
     \vspace{-2mm}
\end{figure*}

\section{Experimental set up}
\vspace{-1mm}
\subsection{Datasets}
We analyze the trade-off between quality and detection performance for two different tasks: summarization and translation. For the summarization task, 1024 contexts are sampled from the test set of XSumm \cite{Narayan2018DontGM}, while for translation 3072 German sentences are sampled from the test set of the XTREME corpus \cite{hu2020xtreme}. 

\subsection{Generative Models}
Two different abstractive summarization systems are used; a BART-based summarization model trained on CNN-daily mail\footnote{\url{https://huggingface.co/facebook/bart-large-cnn}} \cite{lewis2019bart} and Zyphr-7B $\beta$ instruction-tuned  \cite{tunstall2023zephyr} which we prompt to perform summarization. For translation, mBART-large-50 is used, which is a BART model fine-tuned for multilingual translation\footnote{\url{https://huggingface.co/facebook/mbart-large-50-many-to-many-mmt}} \cite{tang2020multilingual}. 

\subsection{Watermarking Methodology}
For each context, the model generates a baseline text without any watermark, and then multiple watermarked texts using various operating points. The watermarking operating points are taken by considering all combinations of green list size $g$ ranging from 0.001 to 0.9 and bias $\delta$ ranging from 0.5 to 8. For summarization, the watermark score is the count of the fraction of green list words in the generated text, while in translation the output texts are grouped in sets of three (to achieve similar expected lengths for detection) and the score is computed for the grouped set. For each operating point, a threshold is chosen to classify watermarked and unwatermarked texts for the maximum $F_{0.5}$ value, which is then used as a detectability metric. $F_{0.5}$ is a weighted harmonic mean of precision and recall, giving more importance to precision than $F_{1}$ to safeguard against false positives. We use the same hashing seed to generate all green-lists, however, Appendix \ref{sec:appendix_seed} shows that consistent results can be observed across random seeds. 


\subsection{Comparative Assessment Set Up}
FLAN-T5 3B \cite{chung2022scaling} is used as the base evaluation LLM, chosen due to its demonstrated pairwise evaluation abilities \cite{liusie2023zero} and good multi-lingual capabilities. As the maximum length of the model is 1024 tokens, if the input prompt exceeds this limit, the end of the context is truncated to fit into the maximum limit, which avoids any of the summaries/translations being truncated. The comparative quality score is taken as the average of all $1024/3072$ samples.



\section{Results}
\label{sec:results}
\vspace{-2mm}



\noindent\textbf{Summarization} Figures \ref{fig:bart} and \ref{fig:zephyr} illustrate the relationship between summary quality and watermark detection performance for BART and Zephyr respectively. A clear trade-off between watermark strength and output quality can be observed for both systems, where strong watermarking degrades quality while weak watermarking maintains quality but yields poor detection performance. The results further suggest that though multiple operating points can yield similar quality-detectability characteristics, the framework provides a simple way to visualize points that achieve a good balance between the two. This can be useful for hyper-parameter selection, e.g. to find the setting where there's minimal quality degradation for a desired $F_{0.5}$ detectability score. Note that the quality scores are upper-bounded near 0.5, consistent with the idea that weak watermarking will enforce little restriction and yield texts of similar quality, while stronger watermarks will restrict generation and therefore yield texts of worse quality. Further, the saturation at $F_{0.5}\!=\!1$ denotes the region where one can perfectly differentiate watermarked texts from unwatermarked texts, albeit often at the cost of large quality degradation.


Additionally, it is observed that different base models can have varying optimal watermarking parameters. Zephyr-7B is much larger than BART (7.2B vs 0.4B parameters) and is likely to have a more accurate underlying task language model. As such, it seems to better deal with the restrictions imposed by watermarking, as seen by the vertical region around the probability of $0.5$ (where minimal quality degradation and good detectability are achieved). Further, in the most extreme settings, Zephyr's average comparative probability drops to 0.44 compared to BART's 0.28. Examples of the generated watermarked text can be seen in Appendix \ref{sec:appendix_txt_examples}. \vspace{2mm}

\begin{figure}[h!]
    \centering
    \includegraphics[width=0.45\textwidth]{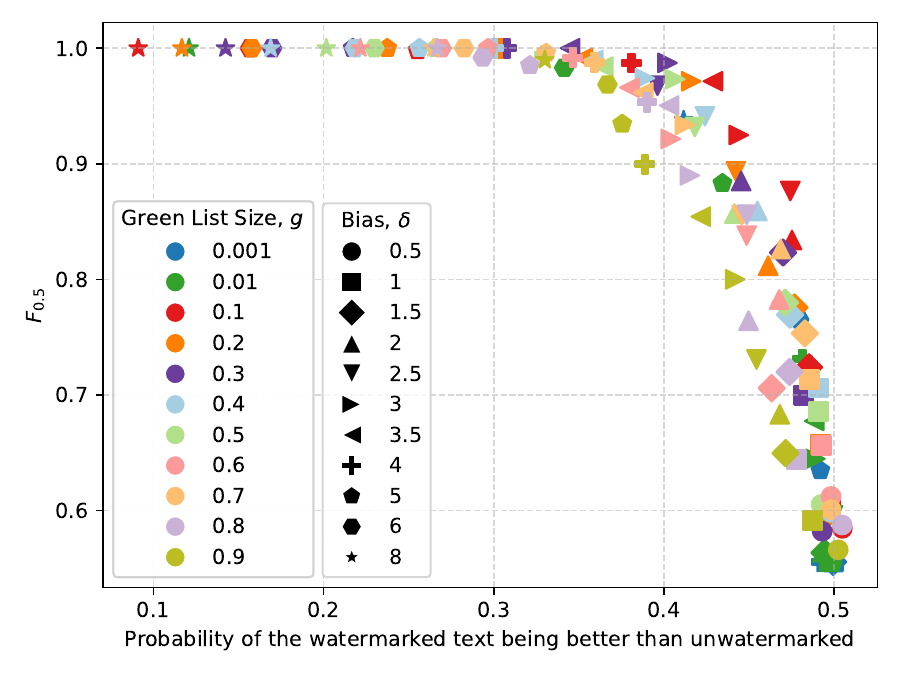}
    \caption{Results for watermarked translations generated with mBART for combinations of green list size $g$ and bias $\delta$.}
    \label{fig:mBART}
    \vspace{-2mm}
\end{figure}

\noindent\textbf{Translation}
We repeat analysis for translation, with Figure \ref{fig:mBART} showing similar quality-detectability characteristics when an mBART system, which translates German sentences to English, is watermarked. The plot shows further evidence of how for weaker models (0.6B parameters supporting 50 languages) strong watermarking can cause a significant drop where the system struggles to maintain quality. Additionally, we can observe that mBART is more sensitive to watermarking parameter settings and that quality is better for small-green list sizes than for larger-green list sizes, even for settings with equivalent detectability performance. 

\begin{figure}[h]
    \centering
    \includegraphics[width=0.45\textwidth]{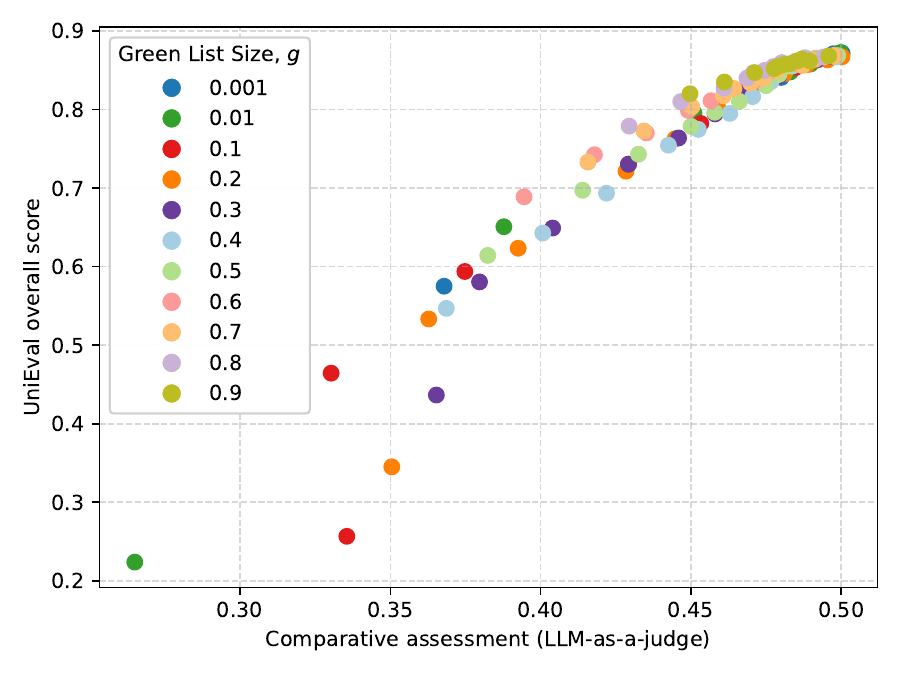}
    \caption{Scatter plot showing correlation between Comparative Assessment and UniEval for BART.}
    \label{fig:eval_comp}
\end{figure}

\noindent\textbf{Suitability of Comparative Assessment} To verify that comparative assessment provides meaningful quality evaluation, we compare the generated quality scores against those from UniEval \cite{zhong2022towards} and COMET\cite{rei2020comet} which both demonstrate strong alignment with human judgment. UniEval is a summary assessment method using a T5-based boolean-answering system trained specifically to assess summaries on coherence, consistency, fluency, and relevance, while COMET is an open-source neural framework for machine translation evaluation. For summarization, comparative assessment has a Spearman correlation of \textbf{0.986} relative to UniEval scores\footnote{scores of the 4 attributes are averaged as an overall score}, while in translation comparative assessment has a Spearman correlation of \textbf{0.988} relative to COMET. Figure \ref{fig:eval_comp} illustrates the relationship between the two quality scores of watermarked summaries generated by BART, with a similar graph for mBART translation shown in Appendix \ref{sec:appendix_comet}). These results highlight that despite being simple and zero-shot, quality assessment via comparative assessment correlates highly with alternative high-performing automatic evaluation approaches that have been tailored to particular tasks. WaterJudge is a clear improvement over more dated metrics such as ROUGE or BLEU, which when used fail to capture the quality-detection trade-off (shown in Appendix \ref{sec:appendix_BLEU}). Further, current popular methods such as perplexity have weaker correlations with UniEval and COMET (\textbf{0.922} for summarization and \textbf{0.940} for translation) and are more difficult to compare between models, where WaterJudge also shows additional promising capabilities, discussed in the next section.

\begin{figure}[h]
    \centering
    \includegraphics[width=0.9\columnwidth]{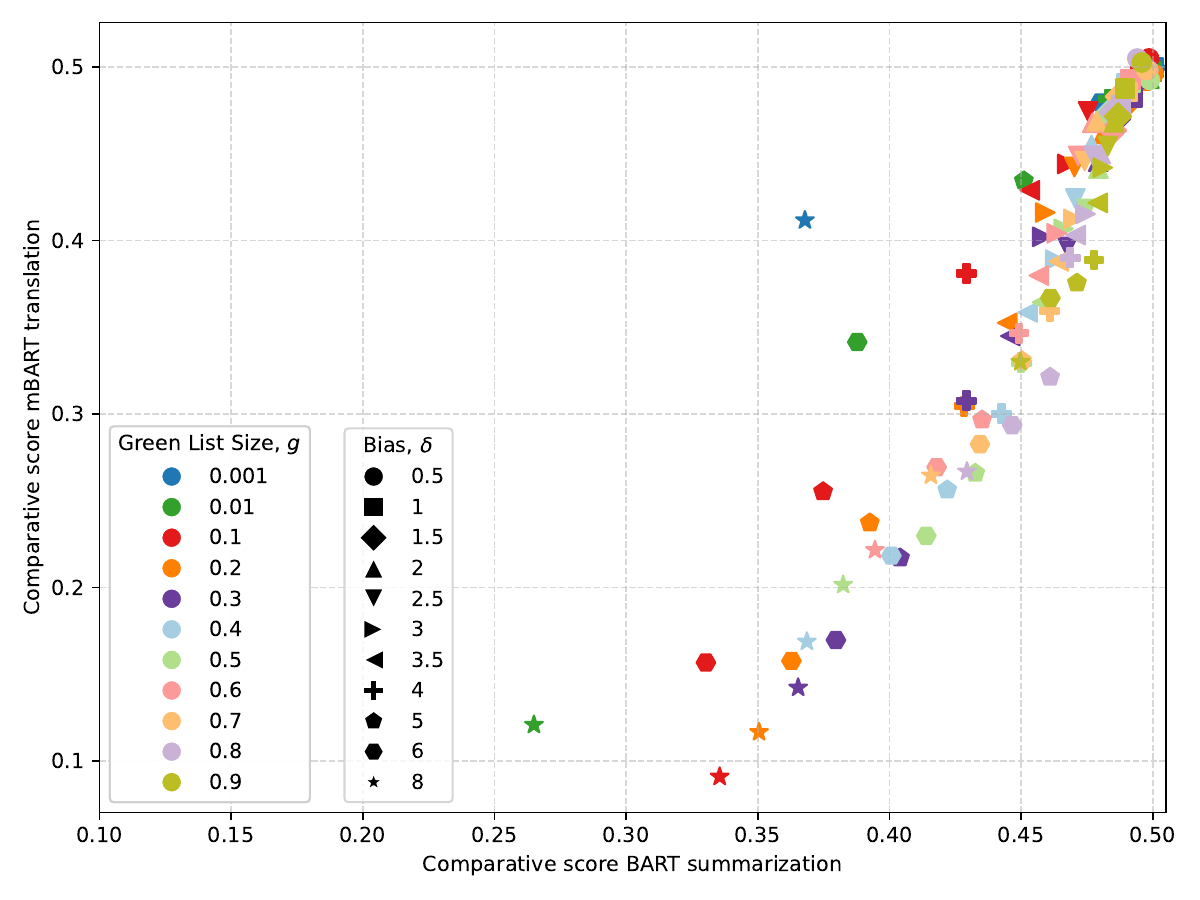}
    \caption{Relationship of watermark settings' comparative assessment quality scores for BART and mBART.}
    \label{fig:model-transfer}
    \vspace{-3mm}
\end{figure}

\begin{figure}[h]
    \centering
    \includegraphics[width=0.9\columnwidth]{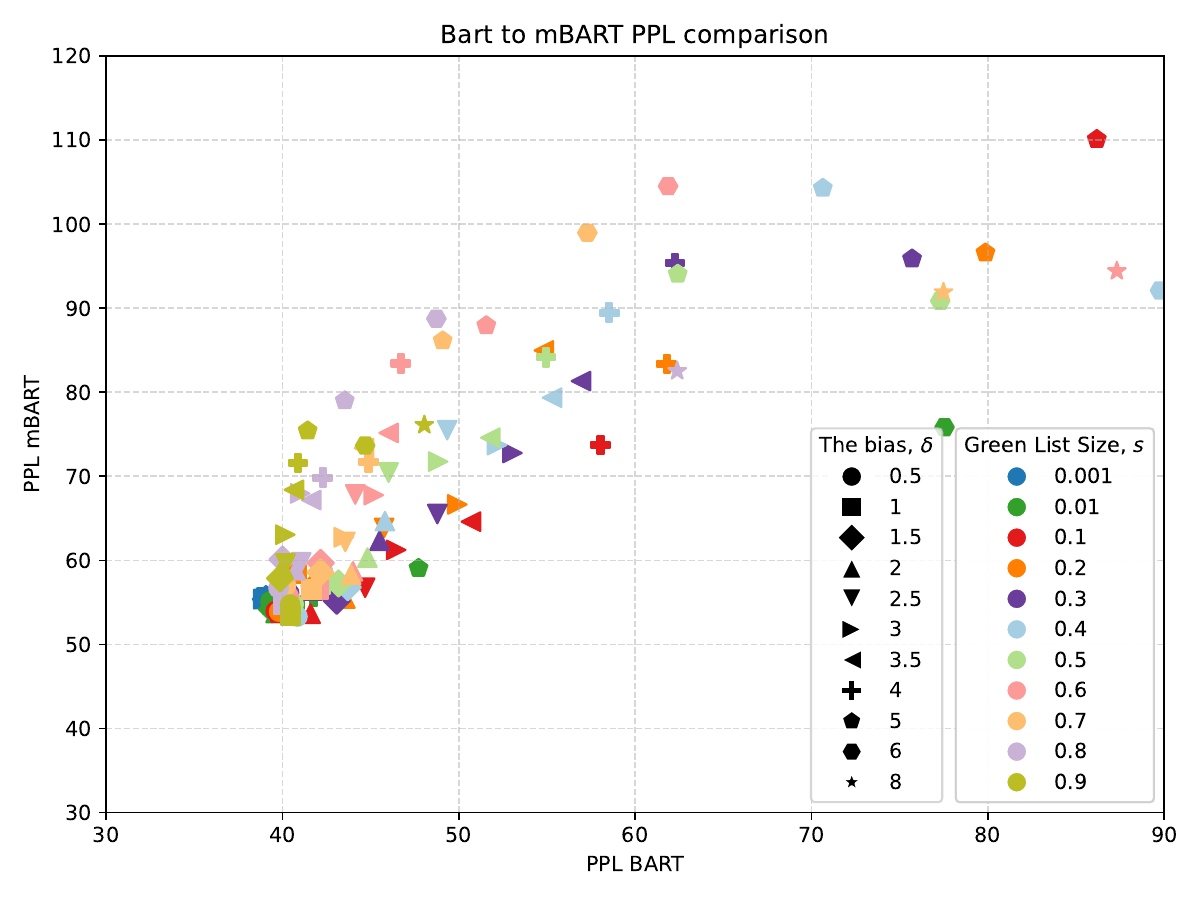}
    \caption{Relationship of watermark settings' perplexity scores for BART and mBART (ignoring the outlier, very high perplexity, points).}
    \label{fig:b_to_mb_ppl}
\end{figure}

\noindent\textbf{Transferability of Settings} As an extension to the current analysis, we consider whether one can avoid doing a full grid search over all watermarking settings and instead transfer settings across different models and tasks. Firstly, it's observed that by looking at the expected quality scores of generated summaries for different operating points on BART and mBART, we observe a Pearson correlation $\rho\!=\!0.927$ (Figure \ref{fig:model-transfer}), while for BART and Zephyr, the correlation is $\rho\!=\!0.826$. The lower correlations for BART-Zephyr can be explained due to the observed truncated linear relationship, where for weak watermarks Zephyr can apply watermarks without causing any quality loss, while for medium watermarks (e.g. $g\!=\!0.5$) there remains a linear degradation to both systems (shown in Figure \ref{fig:b_to_z_qual} in the Appendix). Using perplexity quality scores does not demonstrate strong cross-system correlations, and as shown in Figure \ref{fig:b_to_mb_ppl}, does not demonstrate the linear relationships that are observed with comparative assessment. Therefore perplexity scores may not be effective when considering transferring watermarking performance. 


\begin{figure}[H]
    \centering
    \includegraphics[width=0.9\columnwidth]{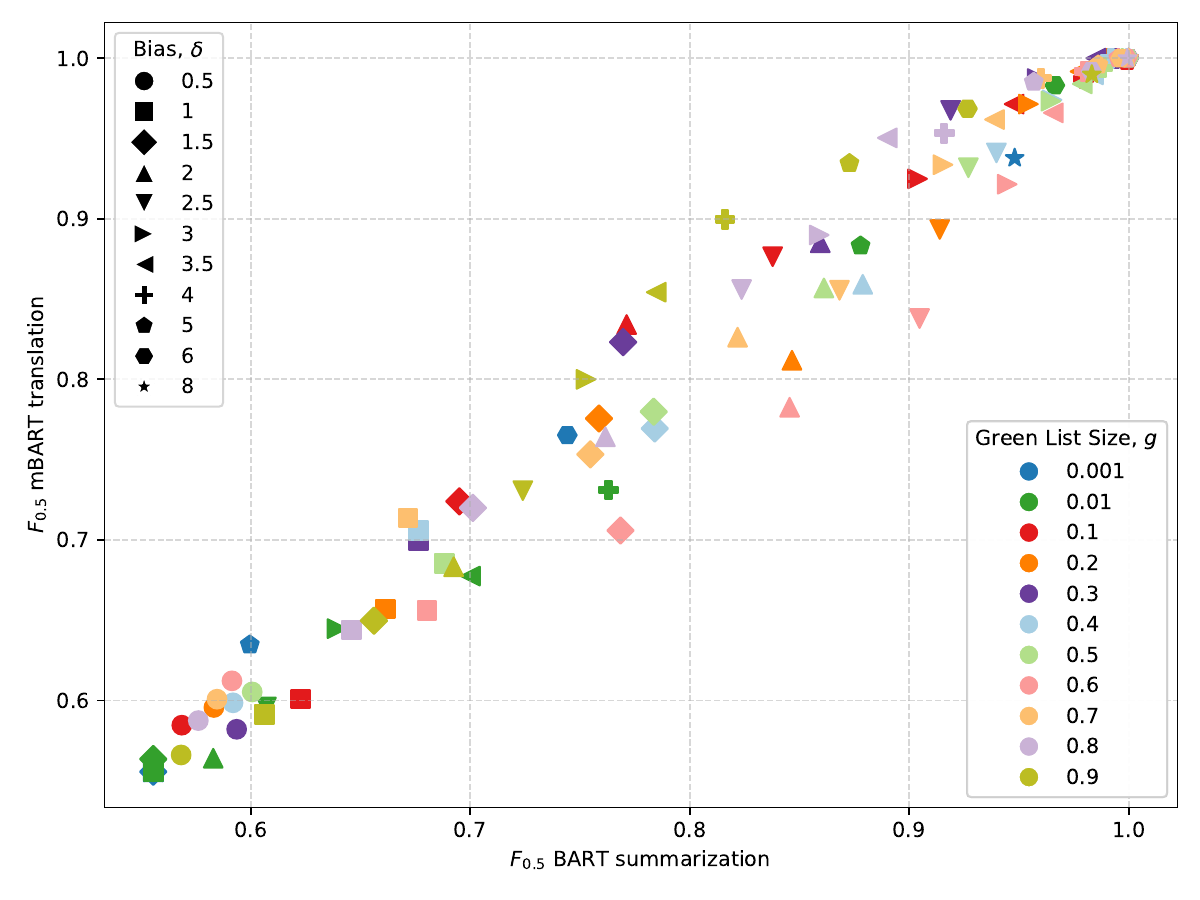}
    \caption{Comparison of watermark settings' detectability scores for BART and mBART.}
    \label{fig:b_to_mb_f05}
    \vspace{-2mm}
\end{figure}

Moreover, the $F_{0.5}$ scores of watermark settings on different models are also highly correlated: BART-Zephyr quality scores have PCC $\rho\!=\!0.986$, while for BART-mBART the PCC is $\rho\!=\!0.990$. Even though this is a cross-task comparison, BART and mBART have a near 1:1 mapping in detectability scores (Figure \ref{fig:b_to_mb_f05}) while Zephyr detectability scores tend to be slightly higher than those from BART (which is mostly due to length mismatches, as discussed in Appendix \ref{sec:appendix_lengths}). The high linear correlations for both quality and detection suggest that WaterJudge can be used to map performance on one model/task to another, which may yield additional predictive abilities for generating the full detectability-quality trade-off curves. Initial examples of the effectiveness of transferring settings across systems are shown in Appendix \ref{sec:appendix_pred}.

\vspace{2mm}

\section{Conclusions}

This paper introduces WaterJudge, a framework for investigating the quality-detection trade-off when watermarking LLMs, enabling easy visualization of various watermarking settings and simple hyper-parameter selection. Comparative Assessment is shown to be a practical metric for measuring quality degradation and improves on currently used evaluation methods in its accuracy and versatility. WaterJudge is also useful in cross-task and cross-model analysis, showing good correlations for both detectability and quality, despite varying characteristics due to model strength.

\section{Limitations}
\vspace{-2mm}
Although LLM evaluation approaches have recently been demonstrated to be effective reference-free evaluation methods, there may be inherent biases such as self-enhancement bias that can impact the robustness of the approach and cause discrepancies in human evaluation. This study could further investigate sensitivity to evaluation prompt sensitivity, or output length, as well as extend to more models, watermarking schemes, and tasks.

\section{Ethical Concerns}
\vspace{-2mm}
Watermark detection performance may not be completely accurate, and false negatives may lead to individuals being unfairly charged for using AIs, when they may have written the text themselves. 

\section{Acknowledgements}
\vspace{-2mm}
This work is supported by Cambridge University Press \& Assessment (CUP\&A), a department of The Chancellor, Masters, and Scholars of the University of Cambridge.

\bibliography{anthology,custom}
\bibliographystyle{acl_natbib}

\appendix
\label{sec:appendix}
\section{Prompts}
\label{sec:appendix_prompts}
\vspace{-2mm}
\begin{table}[h]
    \centering
    \small
    \begin{tabular}{|c|c|}
        \hline
        Use case & Prompt \\
        \Xhline{2pt}
        \multirow{7}{*}{\parbox{2cm}{Zephyr \\ summarization}} & \multirow{7}{*}[-1ex]{\parbox{4cm}{<|system|>\\
You are a tool providing a short text summary.\\
<|user|>\\
Write a short summary of the following text: {context}\\
<|assistant|>}} \\
        &  \\
        &  \\
        &  \\
        &  \\
        &  \\
        &  \\
        &  \\
        \hline
        \multirow{6}{*}{\parbox{2cm}{FLAN-T5 \\ summarization \\ Comparative \\ Assessment}} & \multirow{6}{*}[-1ex]{\parbox{4cm}{Passage: \{passage\}\\ Summary A: \{summary 1\}\\ Summary B: \{summary 2\}\\ Between Summary A and Summary B, which text summarises the passage better?}} \\
        &  \\
        &  \\
        &  \\
        &  \\
        &  \\
        &  \\
        \hline
        \multirow{6}{*}{\parbox{2cm}{FLAN-T5 \\ translation \\ Comparative \\ Assessment}} & \multirow{6}{*}[-1ex]{\parbox{4cm}{Original text: \{context\}\\ Translation A: \{translation 1\}\\ Translation B: \{translation 2\}\\ Between Translation A and Translation B, which is the better translation of original text?}} \\
        &  \\
        &  \\
        &  \\
        &  \\
        &  \\
        &  \\
        \hline
    \end{tabular}
    \caption{prompts used for experiments.}
    \label{tab:prompts}
\end{table}

\noindent For reproducibility, Table \ref{tab:prompts} shows the prompts used for summary generation (using Zephyr 7B $\beta$) and comparative assessment (with FLAN-T5 as the base LLM). For summarization, we evaluated the overall summary quality, as in initial experiments where particular attributes were assessed, the LLM struggled to differentiate between the different attributes with simple prompts (e.g. 'fluency', 'coherence', 'consistency', or 'relevance'). We use a Tesla V100S 32Gb GPU to conduct all experiments. FLAN-T5 Comparative assessment takes 6 minutes to assess each summarization watermark operating point (1024 samples) and 10 minutes to assess each translation operating point (3072 samples). It takes 5 minutes for BART to generate 1024 summaries, 40 minutes for Zephyr-7B $\beta$ to generate 1024 summaries, and MBART 12 minutes to generate 3072 translations.


\vspace{-2mm}
\section{Watermarked texts length}
\label{sec:appendix_lengths}

\vspace{-2mm}

\begin{figure}[H]
    \centering
    \includegraphics[width=0.9\columnwidth]{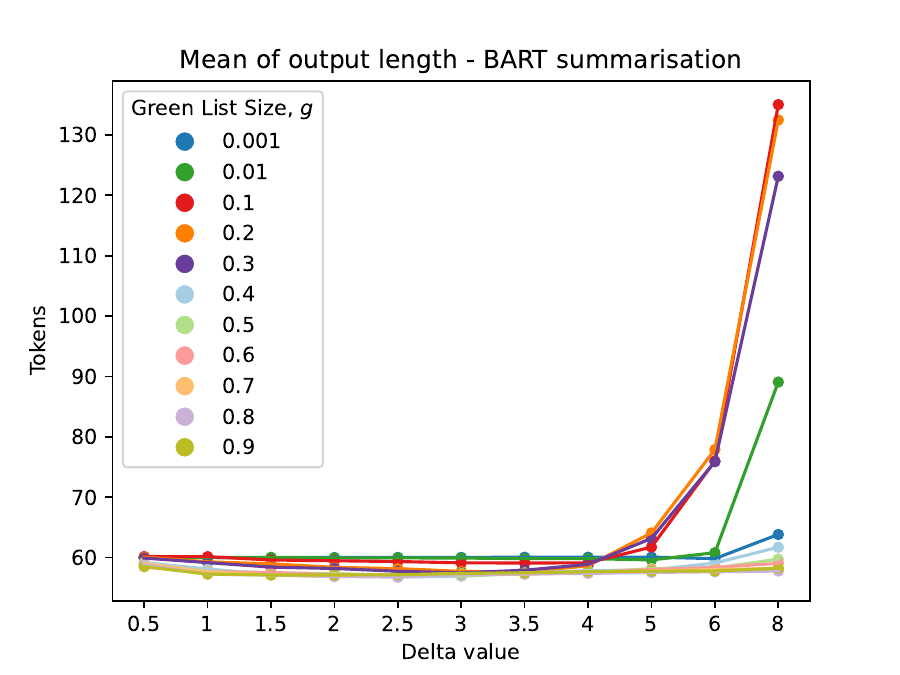}
    \caption{Average length (in tokens) of output BART (summarization) texts for various watermark settings.}
    \label{fig:length-bart}
    \vspace{-4mm}
\end{figure}

\begin{figure}[H]
    \centering
    \includegraphics[width=0.9\columnwidth]{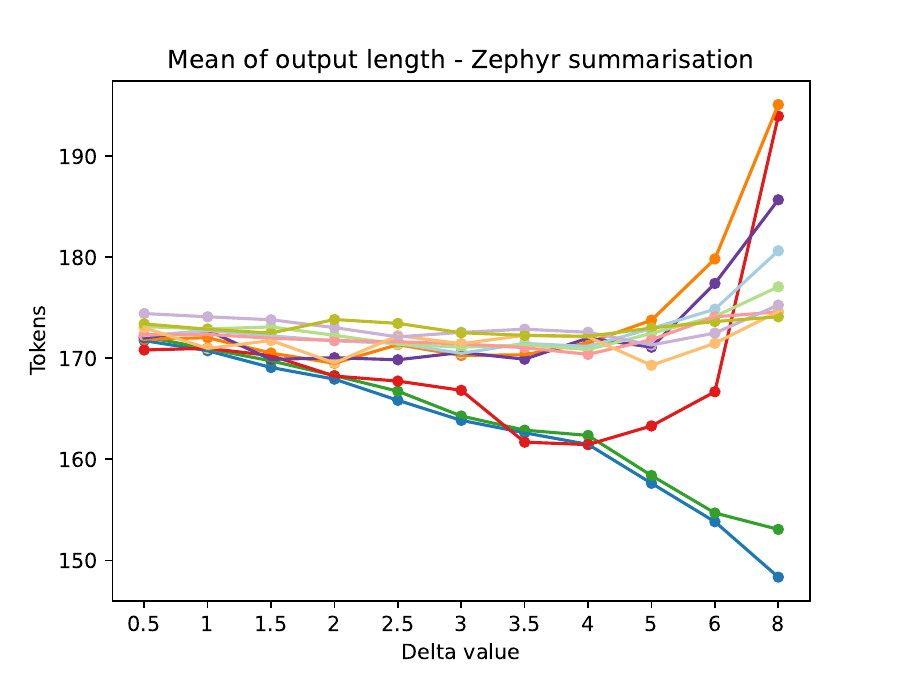}
    \caption{Average length (in tokens) of output Zephyr (summarization) texts for various watermark settings.}   
    \label{fig:length-zephyr}
    \vspace{-4mm}
\end{figure}

\begin{figure}[H]
    \centering
    \includegraphics[width=0.9\columnwidth]{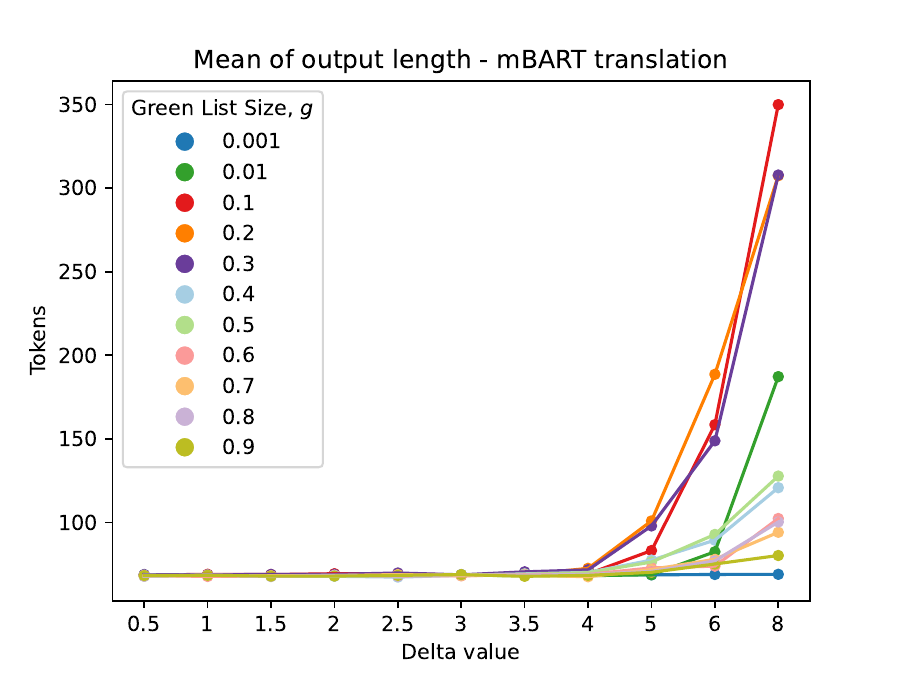}
    \caption{Average length (in tokens) of output mBART (translation) texts for various watermark settings.}
    \label{fig:length-mbart}
\end{figure}

Figures \ref{fig:length-bart}, \ref{fig:length-zephyr} and \ref{fig:length-mbart} show the average lengths (in tokens) of the outputs of the models. BART and mBART were fine-tuned for a specific task and therefore the outputs typically have consistent length (usually 60-80 tokens). Zephyr 7B $\beta$ tends to generate longer summaries with a larger variance in the output lengths. Note that longer texts will typically be easier to detect since having more generated words will reduce the expected variance from the expected fraction of green list words. Therefore when choosing optimal operating points, one should also take the length into account.

\noindent Moreover, the average lengths for models show an issue with simple watermark partitioning into green/red lists: if `</s>' is in the red list of `.', then the outputs tend to grow overly long (since ending the sequence incurs a red-list word). In our experiments, green lists are subsets of larger green lists, and, for both models, the aforementioned issue occurs when $g < 0.4$. High bias $\delta$ texts with $g < 0.4$ are significantly longer than those from other settings, which considerably impacts text quality (though for very small green lists, $g \leq 0.001$, there are more red-list words generated and therefore the sentence may end as expected). For the given randomized seed, Zephyr does not have eos token `</s>' in green list of `.' (for any $g > 0.9$). Hence, the rise is visible for all larger green lists $g \geq 0.1$, but the issue is not as significant as for the other models (due to its better capability to adapt to watermark restrictions). This highlights that this problem is significant when evaluating very strong bias operating points (with generally unusable outputs), but does not otherwise influence evaluation (see Appendix \ref{sec:appendix_seed}).

\section{Baseline Evaluation Metrics}
\label{sec:appendix_BLEU}

\begin{figure}[h]
     \centering
     \begin{subfigure}{0.4\textwidth}
         \includegraphics[width=\linewidth]{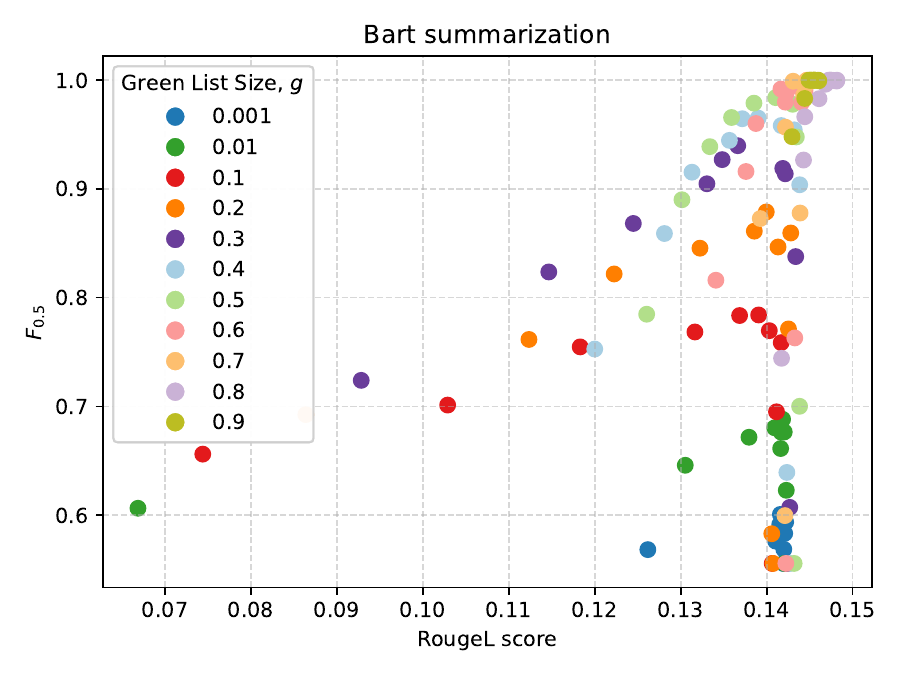}
         \caption{$F_{0.5}$ against RougeL scores for BART summarization.}
         \label{fig:bart-bleu}
     \end{subfigure}
     \hspace{0.05\textwidth}
     \begin{subfigure}{0.4\textwidth}
         \includegraphics[width=\linewidth]{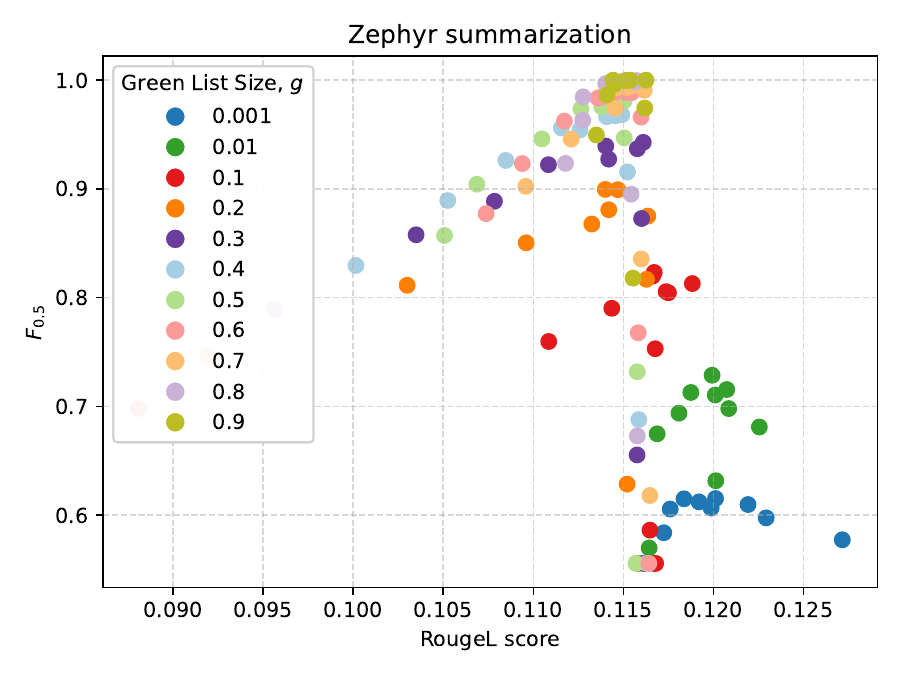}
         \caption{$F_{0.5}$ against RougeL scores for Zephyr summarization.}
         \label{fig:zephyr-bleu}
     \end{subfigure}
     \hspace{0.05\textwidth}
     \begin{subfigure}{0.4\textwidth}
         \includegraphics[width=\linewidth]{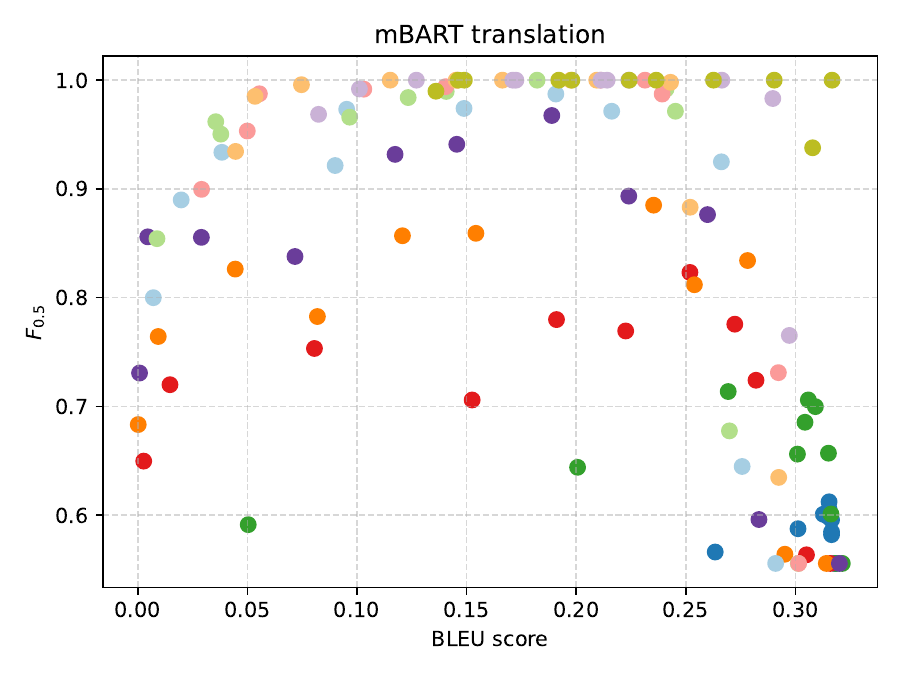}
         \caption{$F_{0.5}$ against BLEU scores for mBART translation.}
         \label{fig:mbart-bleu}
     \end{subfigure}
     \caption{Quality-Detectability trade-off curves for commonly used similarity metrics.}
     \label{fig:appendix_metrics}
\end{figure}

Instead of using comparative assessment to assess the quality degradation, we generate equivalent plots using metrics such as ROUGE or BLEU (against reference summaries/translations), which are standard watermark evaluation metrics. Figure \ref{fig:appendix_metrics} shows that using these evaluation metrics leads to curves that mask the quality-detection tradeoff and provide little insight. The RougeL curves seem to be strongly influenced by summary length (see Figures \ref{fig:length-bart}, \ref{fig:length-zephyr}, \ref{fig:length-mbart}), while the BLEU metric has little explanation. This highlights that the WaterJudge framework requires a capable and effective evaluation approach to capture the quality-detection trade-off and that comparative assessment is a suitable method.

\vspace{-2mm}

\section{Green List Seed Consistency}
\label{sec:appendix_seed}

Results in Appendix \ref{sec:appendix_lengths} suggested that there may be some seed variability, dependent on specific word (or special token) green list bi-grams. To verify the consistency of our results, evaluation for three different seeds is shown in Table \ref{tab:seeds}. The seeds were selected to maximize variability, such that for seed 1 `</s>' is never in the green list of `.', for seed 2 it occurs when $g > 0.4$, and for seed 3 `</s>' is always in the green list of `.' It's observed that even in these settings, there is little impact on the metrics for the main operating points ($\delta$ is 3 or 6), and only when in regions with very heavy watermarks ($\delta$ = 9) are small differences seen. It is worth noting that the length is affected by the seeds, but it's not necessarily negatively received by the Comparative Assessment (in contrast to metrics like Rouge). Due to this designed length bias, seed 1 does on average report slightly higher $F_{0.5}$.

\begin{table}[h]
    \centering
    \small
    \begin{tabular}{cc|ccc|ccc}
        \toprule
        && \multicolumn{3}{c|}{$F_{0.5}$} & \multicolumn{3}{c}{Quality} \\
        $g$ & $\delta$ & 1 & 2 & 3 & 1 & 2 & 3 \\
        \midrule
         0.2   &    3     &  0.90 & 0.86  & 0.88  & 0.44  &  0.44 & 0.44 \\
        \midrule
         0.5   &    3      & 0.90  & 0.89  & 0.86  & 0.44  & 0.44  & 0.43  \\
        \midrule
         0.8   &    3      & 0.76  & 0.73  & 0.72  & 0.46  & 0.45  &  0.45 \\
        \midrule
         0.1   &    6     & 1.00  & 1.0  & 1.0  & 0.28  & 0.29  & 0.29  \\
        \midrule
         0.4   &    6      & 1.00  & 1.0  & 1.0  & 0.35  & 0.34  & 0.33  \\
        \midrule
         0.7   &    6     & 0.99  &  0.99 & 0.99  & 0.39  & 0.40  & 0.39  \\
        \midrule
         0.01   &    9      & 1.0  & 1.0  & 1.0  & 0.15  & 0.12  & 0.12 \\
        \midrule
         0.2   &     9     & 1.0  & 1.0  & 1.0  & 0.16  & 0.15  & 0.16  \\
        \midrule
         0.3   &     9     & 1.0  & 1.0  & 1.0  & 0.23  & 0.19  & 0.20  \\
        \midrule
         0.6   &     9     & 1.0  & 1.0  & 1.0  & 0.31  & 0.30  & 0.28  \\
    \bottomrule
    \end{tabular}
    \caption{Table comparing detectability and quality scores of three additional seeds for various operating points in BART summarization , with good agreement between all seeds.}
    \label{tab:seeds}
\end{table}

\section{Model to model comparison}
\label{sec:appendix_model-to-model}
\begin{figure}[H]
    \centering
    \includegraphics[width=0.9\columnwidth]{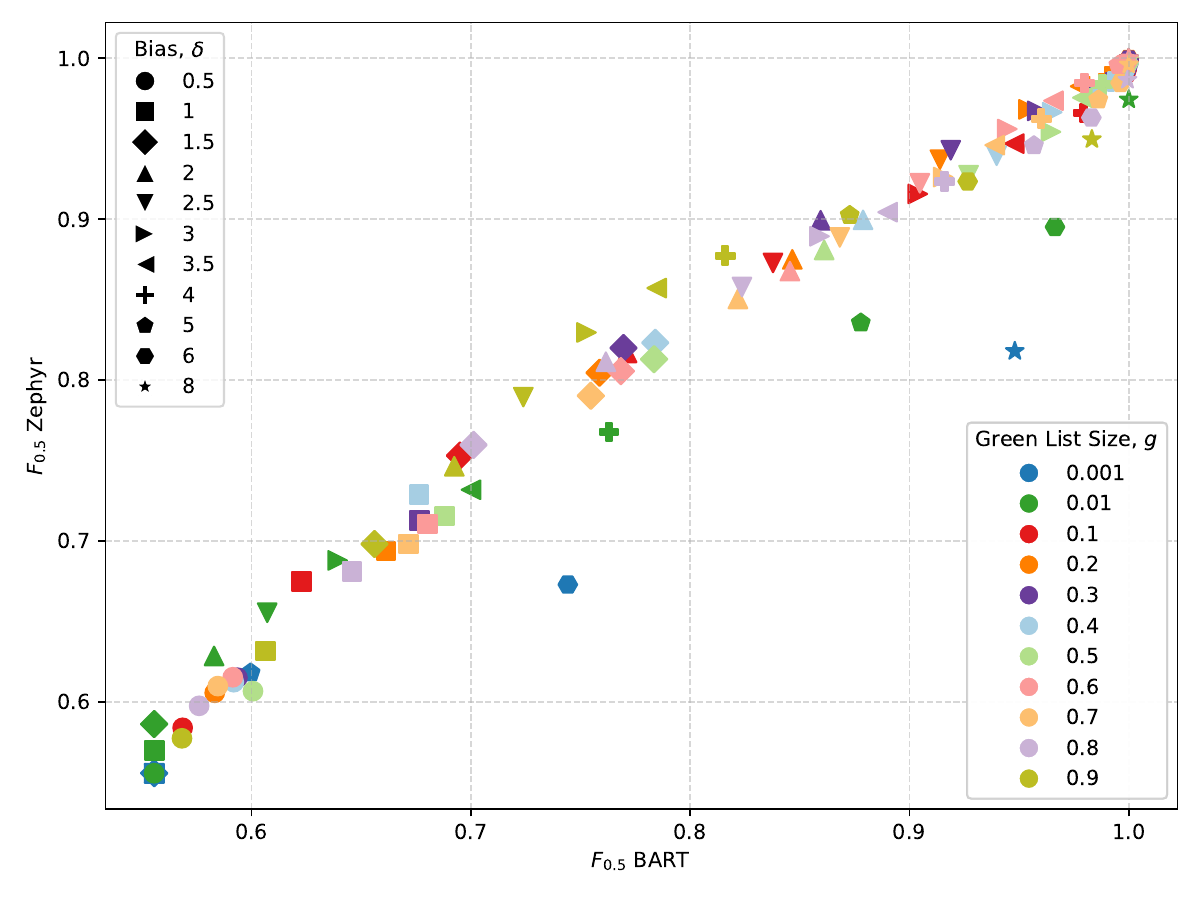}
    \caption{Comparison of watermark settings' detectability scores for BART and Zephyr}
    \label{fig:b_to_z_f05}
    \vspace{-2mm}
\end{figure}

\begin{figure}[H]
    \centering
    \includegraphics[width=0.9\columnwidth]{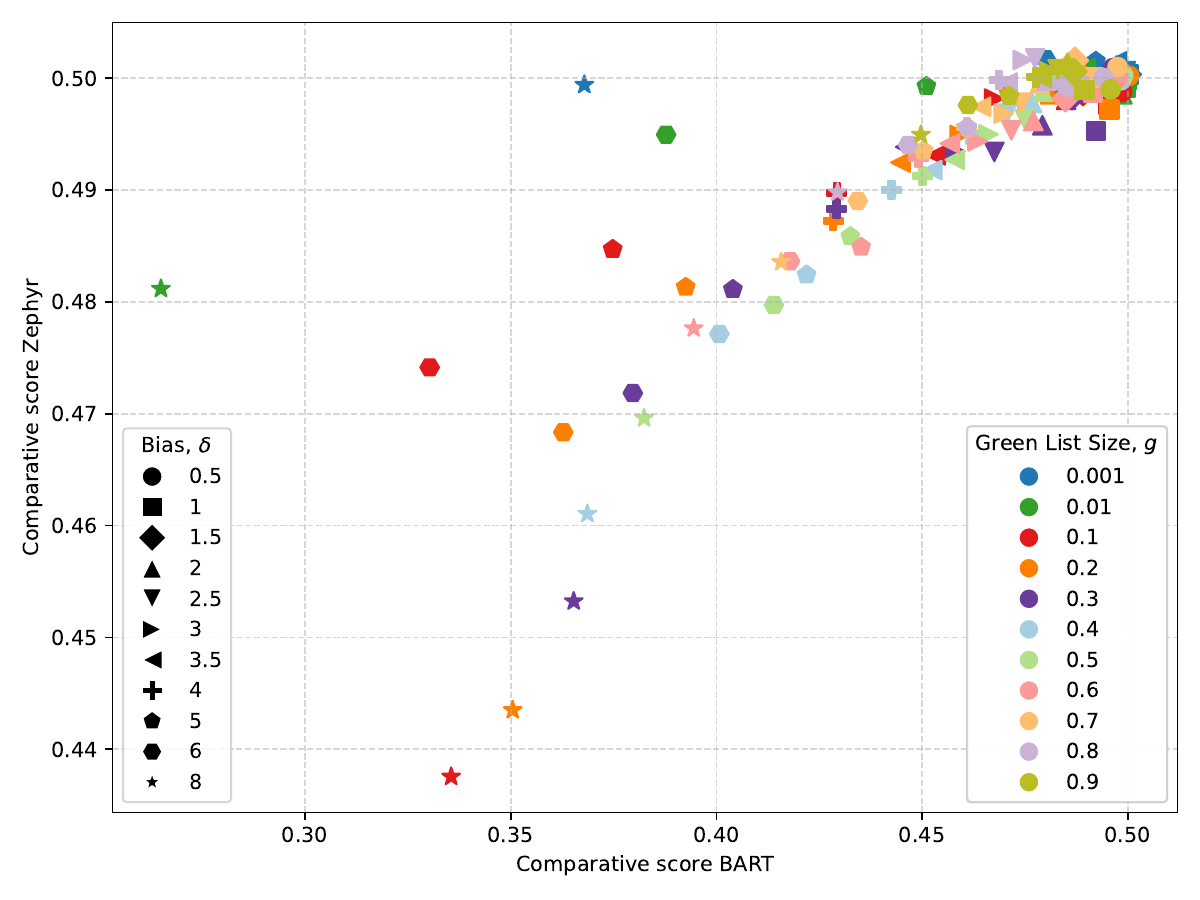}
    \caption{Comparison of watermark settings' quality scores for BART and Zephyr}
    \label{fig:b_to_z_qual}
    \vspace{-4mm}
\end{figure}

\begin{figure}[H]
    \centering
    \includegraphics[width=0.9\columnwidth]{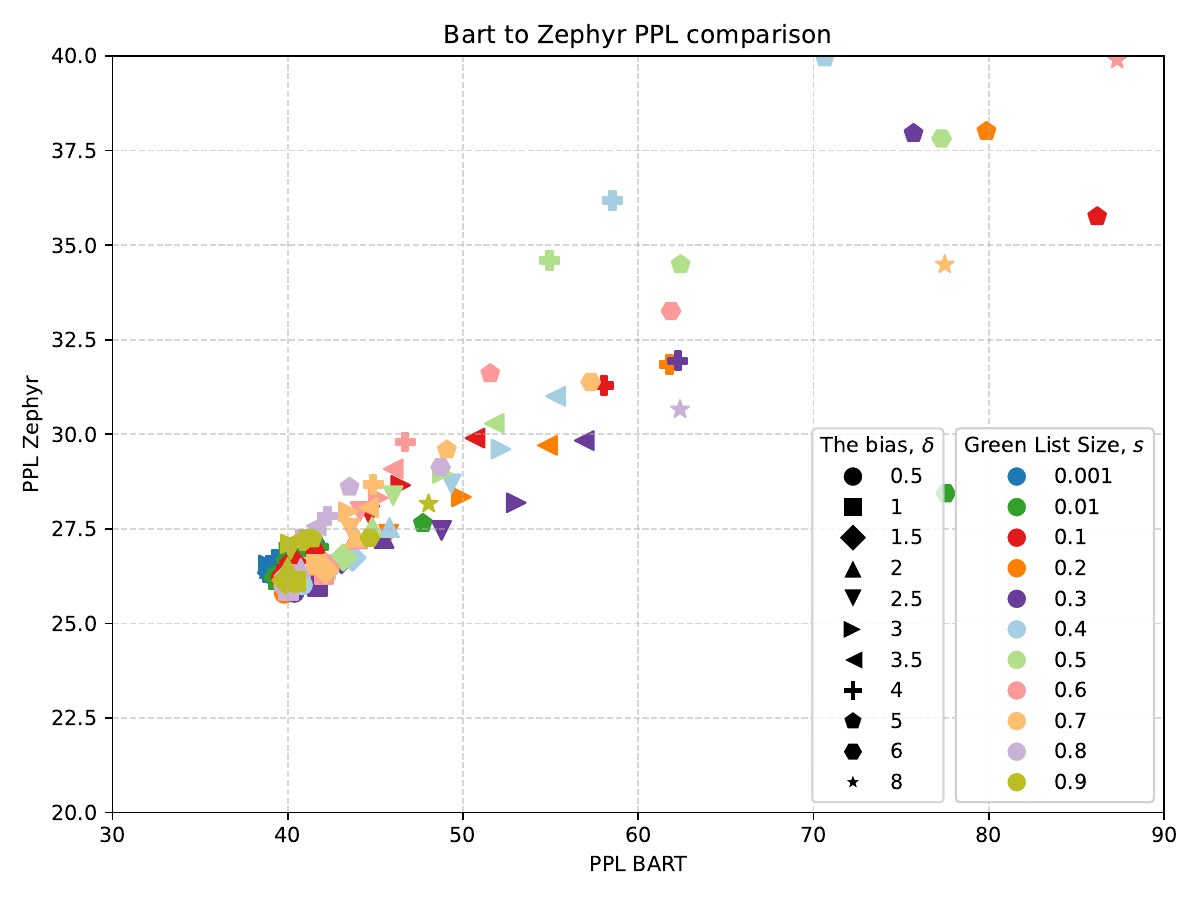}
    \caption{Comparison of watermark settings' PPL for BART and Zephyr (ignoring the outlier, very high perplexity, points).}
    \label{fig:b_to_z_ppl}
\end{figure}

Figures \ref{fig:b_to_z_f05} details the relationship of \textit{detectability} for different watermarking settings for BART-Zephyr, while Figure \ref{fig:b_to_z_qual} shows the equivalent graphs for \textit{quality}. Figure \ref{fig:b_to_z_ppl} shows the BART to Zephyr comparison for Perplexity, where notably most of the points are tightly grouped in a single region. Note that highly hallucinated outputs have extremely high perplexity scores, and so have been cropped out of the plot. Moreover, Figure \ref{fig:b_to_z_qual} suggests that for most models, large green list sizes can yield reasonable detectability with minimal quality degradation, medium green list sizes have predictable and transferable linear degradation, while small/very small green lists have unpredictable behavior and should be avoided.




\section{WaterJudge predicitive capabilities}
\label{sec:appendix_pred}
\begin{figure}[h]
    \centering
    \includegraphics[width=0.8\columnwidth]{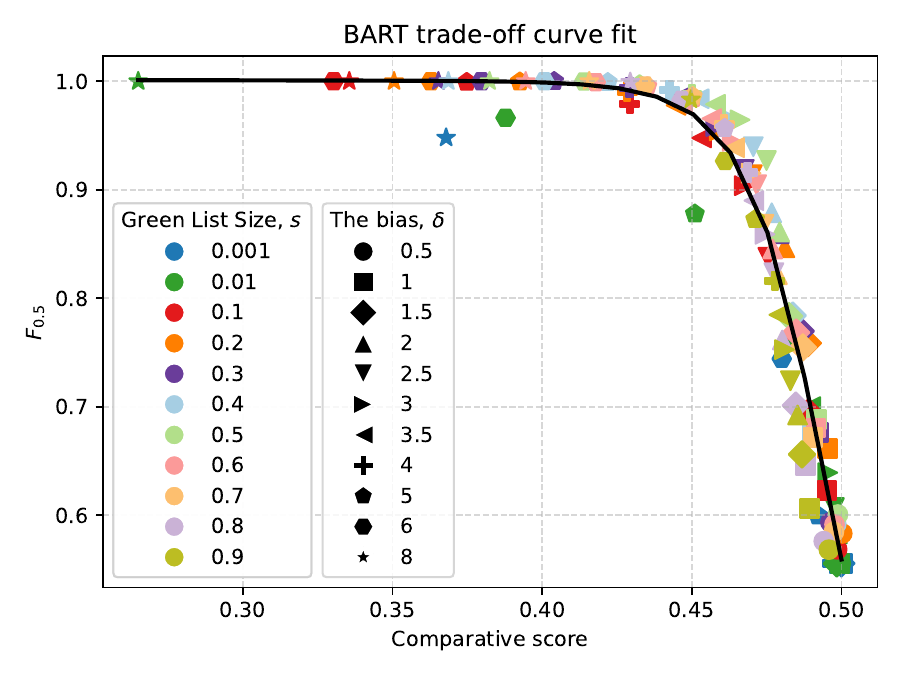}
    \caption{Curve fit to BART operating points graph.}
    \label{fig:bart_fit}
\end{figure}

Appendix \ref{sec:appendix_model-to-model} and Section \ref{sec:results} demonstrated that both Zephyr-BART and mBART-BART have consistent and linear relationships between quality and predictive scores. Detectability is nearly equivalent across different watermarking settings, starting from $F_{0.5} = 0.5$ and linearly increasing to $F_{0.5} = 1$. Alternately, the Quality comparisons can be broken into three regions: for weak watermark settings, large models (like Zephyr) can maintain quality, for medium strength watermarks (e.g. $g\!=\!0.5$) there is a linear degradation for all systems, while strong watermarks can lead to meaningless output texts and low transferability (deviations from the trend). Meaningful regions of both of these curves can be estimated well with a two-parameter function (such as truncated at the top linear function), which enables a transformation of watermark performance from one system to another while using only a few tested operating points. To achieve the fitting, a parameterized hyperbolic tangent was fitted to the BART quality-detection curve, as shown in Figure \ref{fig:bart_fit}, by minimizing the average perpendicular Mahalanobis distance of the operating points from the curve. This has been done to get a smooth baseline function capturing the whitened data shape. 



\begin{figure}[h!]
     \centering
     \begin{subfigure}{0.35\textwidth}
         \includegraphics[width=\linewidth]{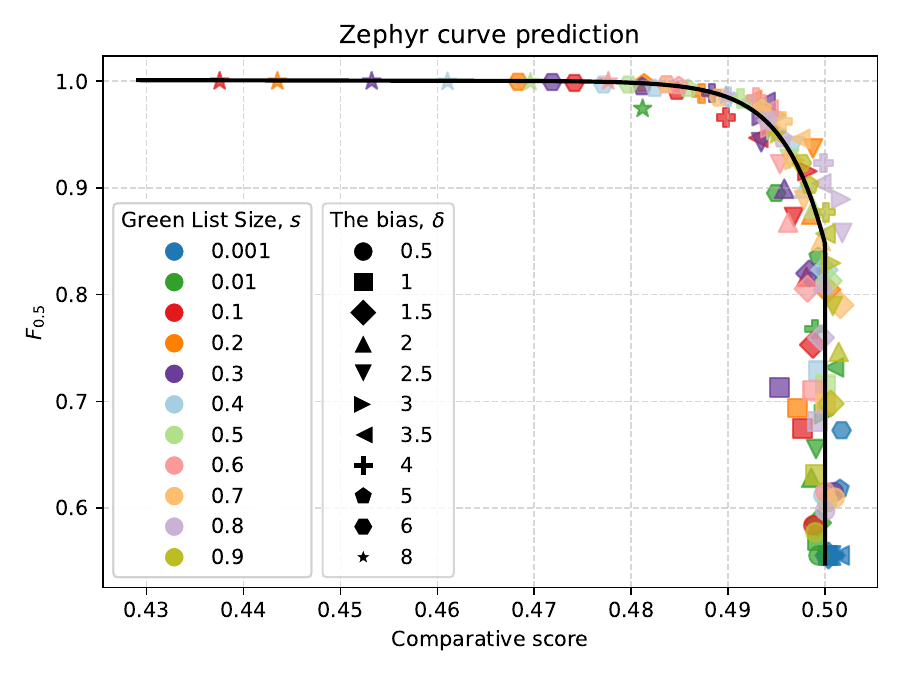}
         \caption{Zephyr}
         \label{fig:zephyr-pred}
     \end{subfigure}
     \hspace{0.05\textwidth}
     \begin{subfigure}{0.35\textwidth}
         \includegraphics[width=\linewidth]{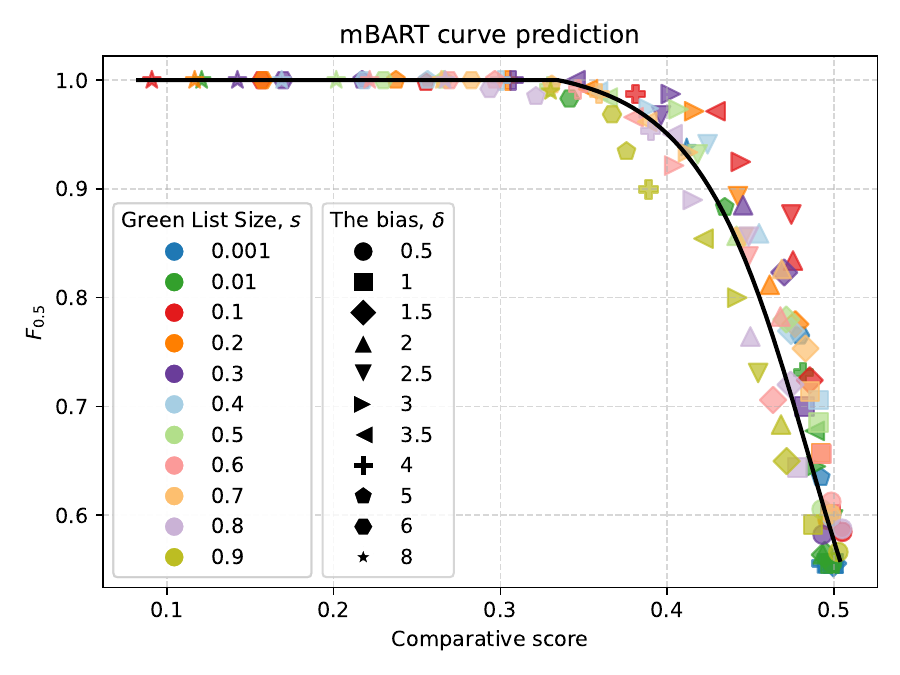}
         \caption{mBART}
         \label{fig:mbart-pred}
     \end{subfigure}
     \hspace{0.05\textwidth}
     \begin{subfigure}{0.35\textwidth}
         \includegraphics[width=\linewidth]{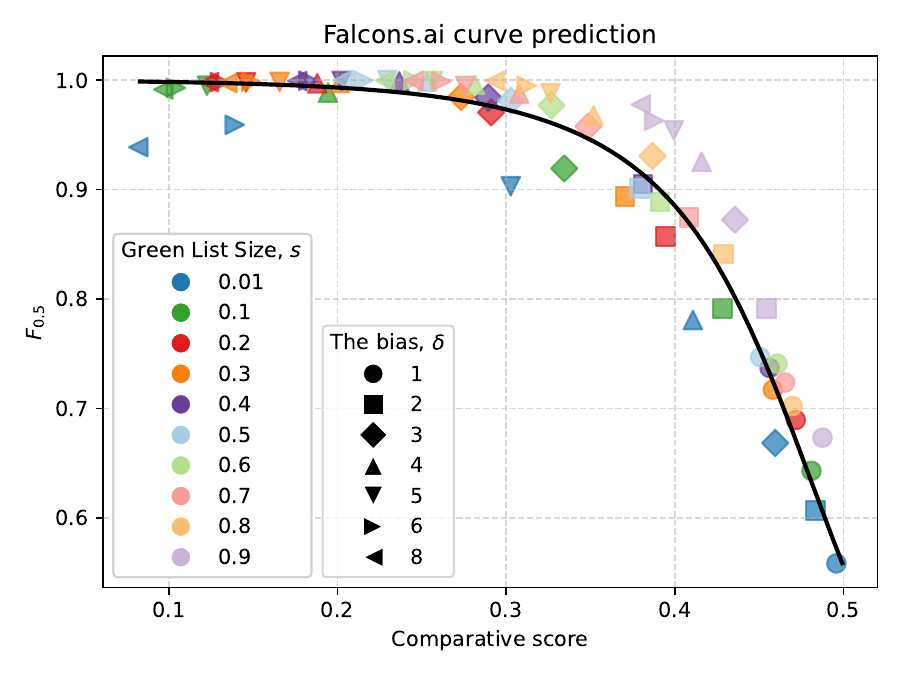}
         \caption{Falcons.ai}
         \label{fig:falcons-pred}
     \end{subfigure}
     \caption{Estimated LLM Quality-Detectability trade-off curves from model-to-model comparisons.}
     \label{fig:appendix_pred}
\end{figure}

\noindent By fitting truncated linear functions to the relationships such as Figures \ref{fig:b_to_z_f05}-\ref{fig:b_to_z_qual}, one can transform a baseline curve to achieve predicted fits, as shown in Figure \ref{fig:appendix_pred}. These 'predicted' shapes are achieved with significantly fewer points and avoid the grid search, which can be useful when attempting to determine whether an effective watermark setting exists for the new system, and also enable fast and thorough testing across hyperparameters and models.


\section{COMET Comparison}
\label{sec:appendix_comet}

\begin{figure}[h!]
    \centering
    \includegraphics[width=0.4\textwidth]{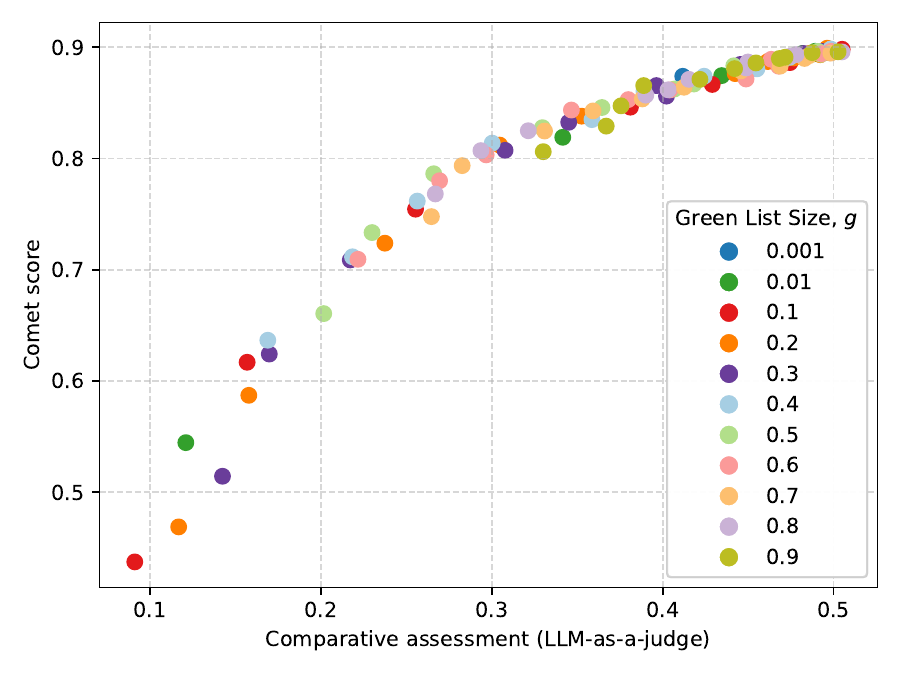}
    \caption{Correlation between Comparative Assessment and COMET score for mBART translation.}
    \label{fig:eval_comparison_comet.pdf}
\end{figure}

Figure \ref{fig:eval_comparison_comet.pdf} shows how Comparative Assessment for translation is also strongly correlated with the most recent automatic translation evaluation metric: COMET. The Spearman correlation of quality scores is \textbf{0.988}.

\onecolumn
\section{Examples of watermarked texts}
\label{sec:appendix_txt_examples}
Table \ref{sec:appendix_txt_examples} provides examples of possible output texts under various operating points for the two summarization experiments for a sampled article. The texts achieve various detectability and show varied degradation in text quality.

\begin{table*}[h]
  \centering
  \small 
  \begin{tabular}{c|c}
    \toprule
    \multirow{7}{*}{Context} & \multirow{7}{*}{\parbox{11cm}{The 41-year-old reached the semi-final for the first time since he last lifted the trophy in 2011 with a 13-6 win over Kyren Wilson.
The world number six will now face the winner of Stephen Maguire's match against Barry Hawkins for a place in Sunday's final.
"I believe I can [win], and that's a great feeling to have," Higgins said.
"I was close last year, Alan [McManus] pipped me [in the quarter-final]," he added.
"A couple of years after 2011, I'd been losing, my form wasn't great and you don't think you're [continues]}} \\
    & \\
    & \\
    & \\
    & \\
    & \\
    & \\
    \Xhline{2pt}
    \multirow{4}{*}{\parbox{3cm}{Unwatermarked BART}} & \multirow{4}{*}{\parbox{11cm}{Higgins beat Kyren Wilson 13-6 to reach the semi-final of the World Championship. The Scot will face the winner of Stephen Maguire's match against Barry Hawkins. The 41-year-old believes he can win the tournament for the first time since 2011.}}\\
     & \\
     & \\
     & \\
    \hline
    \multirow{4}{*}{\parbox{3cm}{$g = 0.7, \delta = 6$ \\ $F_{0.5} = 0.996$ \\ Quality score $0.434$}} & \multirow{4}{*}{\parbox{11cm}{Higgins beat Kyren Wilson 13-6 in the quarter-finals at the World Cup. He now awaits the winner of Stephen Maguire's match against Barry Hawkins. It is the first time the 41-year-old Scot is in the semi-final since he lifted the trophy in 2011.}}\\
     & \\
     & \\
     & \\
    \hline
    \multirow{4}{*}{\parbox{3cm}{$g = 0.5, \delta = 4$ \\ $F_{0.5} = 0.988$ \\ Quality score $0.450$}} & \multirow{4}{*}{\parbox{11cm}{Scottishman Higgins is into the semi-final for the first time since 2011. He beat Kyren Wilson 13-6 in his quarter-final on Wednesday. He now faces the winner of Stephen Maguire and Barry Hawkins. He says he has more self-confidence heading to final.}}\\
     & \\
     & \\
     & \\
    \hline
    \multirow{5}{*}{\parbox{3cm}{$g = 0.1, \delta = 8$ \\ $F_{0.5} = 1.00$ \\ Quality score $0.336$}} & \multirow{5}{*}{\parbox{11cm}{Bobbyiggins into first semifinal since picking off David Higgins 11 years ago. 41-year-old Scot defeated World 16's KYRN WEIRD 12 4 8 3 3 3 8 1 this night at  BAVCA PIA Stadium. Beats Ian Pinn are first-round bye; winner of Stephen MagUires takes part Saturday afternoon and it goes 6 vs 6 - winner to meet for the tournament [continues]}}\\
     & \\
     & \\
     & \\
     & \\
    \Xhline{2pt}
    \multirow{5}{*}{\parbox{3cm}{Unwatermarked Zephyr}} & \multirow{5}{*}{\parbox{11cm}{John Higgins, aged 41, advanced to the snooker World Championship semi-finals for the first time in a decade with a 13-6 victory over Kyren Wilson. Higgins will face either Stephen Maguire or Barry Hawkins for a place in the final, and believes he can win the tournament. The Scot has won all three sessions against Wilson and [continues]}}\\
     & \\
     & \\
     & \\
     & \\
    \hline
    \multirow{5}{*}{\parbox{3cm}{$g = 0.5, \delta = 4$ \\ $F_{0.5} = 0.984$ \\ Quality score $0.491$}} & \multirow{5}{*}{\parbox{11cm}{John Higgins, 41, progressed to his seventh World Snooker Championship semi-final, beating Kyren Wilson 13-6. Higgins, ranked sixth, aims for his fourth title after last lifting the trophy in 2011. The Scottish player faces either Stephen Maguire or Barry Hawkins in the last four, with Higgins preferring a Maguire match as a fellow [continues]}}\\
     & \\
     & \\
     & \\
     & \\
    \hline
    \multirow{6}{*}{\parbox{3cm}{$g = 0.1, \delta = 8$ \\ $F_{0.5} = 1.00$ \\ Quality score $0.438$}} &\multirow{6}{*}{\parbox{11cm}{John Higgins, aged 41, advanced to the world snooker championship semi-finals for the first time since lifting the trophy in 2011. In the quarter-finals, Higgins defeated Kyren Wilson in three sessions by a scoreline of 13-6. In the semi-finals, Higgins will confront either Stephen Maguire or Barry Hawkins for a position in the final on Sunday. Reflecting after his match against Wilson, Higgins stated that he "believ(ed) (he) [continues]}}\\
     &\\
     &\\
     &\\
     &\\
     &\\
     \bottomrule
  \end{tabular}
  \caption{Examples of watermarked and unwatermarked outputs for a few chosen operating points for summarization. It may be noted that in the strong watermarking region, the model begins `hallucinating' or repeating text continuously. This yields long, unusable texts with high density of green list words, making the operating points a bit less meaningful for analysis. This is more difficult to trigger for larger models like Zephyr.}
  \label{tab:sample_texts_table}
\end{table*}

\end{document}